# Learning is planning: near Bayes-optimal reinforcement learning via Monte-Carlo tree search

**John Asmuth**
Department of Computer Science
Rutgers University
Piscataway, NJ 08854

**Michael Littman**
Department of Computer Science
Rutgers University
Piscataway, NJ 08854

## Abstract

Bayes-optimal behavior, while well-defined, is often difficult to achieve. Recent advances in the use of Monte-Carlo tree search (MCTS) have shown that it is possible to act near-optimally in Markov Decision Processes (MDPs) with very large or infinite state spaces. Bayes-optimal behavior in an unknown MDP is equivalent to optimal behavior in the known belief-space MDP, although the size of this belief-space MDP grows exponentially with the amount of history retained, and is potentially infinite. We show how an agent can use one particular MCTS algorithm, Forward Search Sparse Sampling (FSSS), in an efficient way to act nearly Bayes-optimally for all but a polynomial number of steps, assuming that FSSS can be used to act efficiently in any possible underlying MDP.

## 1 Introduction

In reinforcement learning (RL), a central issue is the exploration/exploitation tradeoff (Sutton & Barto, 1998). Simply put, this dilemma refers to the balance between acting optimally according to the information that you have (exploitation) and acting potentially sub-optimally so as to improve the quality of your information (exploration).

The classical approach to this issue considers the "general" model, and makes its guarantees (if any exist) about an algorithm's behavior on any possible model that satisfies some constraints. A maximum likelihood estimate (MLE) is used with a promise that, if the correct exploration steps are taken, the resulting model is close to the truth.

This frequentist approach to optimal behavior is effective in many scenarios, and has the added benefit that the MLE is often easy to compute. However, providing guarantees for a "general" MDP can cause *over-exploration* beyond what is needed to ensure optimal or near-optimal behavior.

An effective approach to limiting over-exploration is to constrain the model to a class that is easier to learn. For example, if we know that the model dynamics can be treated as a separate problem for any one of a number of state features, we can more efficiently do factor learning (Strehl et al., 2007). Or, if we know that certain groups of states have identical dynamics, we can learn the group dynamics by using relocatable action models (Leffler et al., 2007). The downside to this strategy is that, in most cases, an entirely new algorithm must be invented to take advantage of new constraints.

The Bayesian approach to model estimation in RL (Wilson et al., 2007) introduces the use of model priors, and much of the innovation required moves from algorithm design to prior engineering and inference. While neither of these new issues are trivial, they extend beyond reinforcement learning and general Bayesian techniques from outside of the field will apply, broadening the palette of possible techniques that can be brought to bear.

One effect of introducing Bayesian ideas to reinforcement learning is the concept of *Bayes-optimal* behavior, which chooses actions that maximize expected reward as a function of belief-state. If an agent achieves Bayes-optimal behavior, it circumvents the exploration/exploitation dilemma; information gathering options are built into the definition. As a result, two major aspects of an RL algorithm, *learning* and *planning*, are unified and become simply *planning*.

Unfortunately, the general problem of computing Bayes-optimal behavior is intractable. *Near* Bayes-optimal behavior, where an agent is approximately Bayes-optimal for all but a small number of steps, has been achieved for Dirichlet priors (Kolter & Ng, 2009) and some more general priors (Sorg et al., 2010).

We will show how an existing planning algorithm, **FSSS**, can be modified to achieve near Bayes-optimal behavior for a very general class of models and model priors.

## 2 Background

Reinforcement learning is a framework for sequential decision-making problems where the dynamics of the environment are not known in advance. In RL, an agent is supplied with an *observation*, to which it responds with an *action*. In response to every action taken, the environment returns a new observation as well as a numerical *reward* signal. The agent's goal is to maximize the total sum of all rewards over time.

Formally, an environment is described as a Markov Decision Process (Puterman, 1994), or MDP. An MDP is the tuple $M = \langle S, A, T, R, \gamma \rangle$, where $S$ is the set of possible states, $A$ is the set of available actions, $T : S \times A \to \Pi(S)$[1] is the transition function, $R : S \times A \to \Pi(\Re)$ is the reward function, and $\gamma$ is a discount factor.

A policy $\pi : S \to A$ describes what action an agent takes in a given state. The optimal policy for some MDP $M$ is defined as follows:

$$\begin{aligned} \pi^* &= \mathop{\mathrm{argmax}}_{a} Q(s,a) \\ Q(s,a) &= E[R(s,a)] + \gamma E_{s' \sim T(s,a)}[V(s')] \\ V(s) &= \max_a Q(s,a). \end{aligned}$$

For any given MDP prior $\phi(M)$, a corresponding belief-MDP $m_\phi$ can be constructed by considering the set of all possible histories $H$, which consists of every finite sequence of steps in the environment. This is also called a Bayes-adaptive MDP (Duff, 2003). The state space of the belief-MDP $\mathcal{S} = S \times H$ pairs "real" states with histories. The action space $A$ remains unchanged. Since the belief-state includes history, next states in the transition function are constrained to belief-states that extend the previous belief-state with the last occurring transition and reward. Because states are completely observable, histories can be summarized by counts of individual state transitions. It is easiest to express the new transition and reward functions as a single joint distribution $T\text{-}R_\phi(\langle s, h\rangle, a) = \Pi(\langle s', h \cup (s,a,s',r)\rangle, r)$.

When $T\text{-}R_\phi$ is constructed in such a way that the likelihood $P(\langle s', h \cup (s,a,s',r)\rangle, r | \langle s,h\rangle, a) = \int_M P(s', r | s, a, M) \phi(M|h) dM$, the optimal policy in $m_{\phi|h}$ corresponds to the Bayes-optimal policy, given the MDP prior $\phi$ (Duff, 2003). In many cases, transition likelihoods are easy to compute or sample from. For example, the **Flat-Dirichlet-Multinomial** (Poupart et al., 2006), or **FDM**,

$$\begin{aligned} \theta_{s,a} &\sim Dir(\alpha) \\ T(s,a) &\sim Mult(\theta_{s,a}), \end{aligned}$$

---
[1] Here, we use the notation that $\Pi(X)$ is the set of probability distributions over the set $X$.

holds that the next-state distributions for each state-action pair are i.i.d. with a Dirichlet prior. In this case, $P(s'|s,a,h) \propto C_{s,a}(s') + \alpha_{s'}$, where $C_{s,a}(s')$ is the number of times a transition to $s'$ has been observed when action $a$ was taken in state $s$, and $\alpha_{s'}$ is the element in the hyperparameter vector $\alpha$ corresponding to $s'$. Note that a contribution of the current paper is to handle general distributions beyond Dirichlet priors.

The obvious intractability problem comes from the size of the belief-MDP's state space $\mathcal{S}$, which grows exponentially with the length of the amount of history to be considered. Even for a very small MDP, the corresponding belief-MDP quickly becomes unmanageable. The use of many standard planning techniques, such as value iteration or policy iteration (Sutton & Barto, 1998; Puterman, 1994), becomes impossible with an infinite horizon domain.

*Exact Bayes-optimal* behavior is impossible to compute efficintly for general MDPs and priors (including **FDM**), so we must rely upon approximations. *Approximate Bayes-optimal* behavior, in which the value of each action selected is within $\epsilon$ of the value of the exact Bayes-optimal action, is easier to achieve than exact Bayes-optimal behavior, but is still intractable. *Near Bayes-optimal* behavior, which is approximately Bayes-optimal for all but a polynomial number of steps, can be efficiently achieved in many situations and is the focus of our work.

## 3 Related Work

The approach we propose in this paper has roots in a number of existing algorithms, which we sketch next.

**Bayesian Exploration/Exploitation Trade-off in LEarning** (Poupart et al., 2006), or **BEETLE**, is an algorithm that uses the belief-MDP formulation of Bayesian RL in order to achieve approximately Bayes-optimal behavior. It uses the **FDM** prior and known properties of the value function to calculate an approximation over all states.

**Bayesian Exploration Bonus** (Kolter & Ng, 2009), or **BEB**, is a near Bayes-optimal algorithm for the **FDM** prior and known rewards. It acts greedily according to the maximum likelihood estimation of the MDP, but uses the alternate reward function $R_{\text{BEB}}(s,a) = R(s,a) + \beta/(1+n(s,a))$, where $\beta$ is a domain-dependent constant and $n(s,a)$ is the number of times action $a$ has been performed in state $s$. This reward-supplement strategy is also used in PAC-MDP (Kakade, 2003) approaches (Strehl & Littman, 2008), and biases an agent toward states in which it has less experience. The particular bonus used by **BEB** will cause the agent to explore enough to be near Bayes-optimal without the over-exploring seen in PAC-MDP algorithms. There are ways to use posterior variance to create reward bonuses for more general Bayesian priors (Sorg et al., 2010).

**Bayesian Dynamic Programming** (Strens, 2000), or **Bayesian-DP**, and **Best Of Sampled Set** (Asmuth et al., 2009), or **BOSS**, are two examples of Bayesian RL algorithms that can work with a much more flexible prior. They both use samples from the MDP posterior. **Bayesian-DP** will sample an MDP at the beginning of the experiment, and will resample the model when the current one has been in use for some threshold number of steps. It acts greedily according to the most recent sample. **BOSS** will sample $C$ models every time a particular state-action pair has been tried $B$ times, and then combine all $C$ models in such a way as to have each state's value be at least as great as that state's value in each of the sampled MDPs. Both algorithms rely on uncertainty in the posterior causing variance in the samples: variance in the samples causes optimistic value estimates, which in turn drive the agent to visit under-explored parts of the state space.

**Sparse Sampling** (Kearns et al., 1999) works by recursively expanding a full search tree up to a certain depth $d$. At the root, each of the $A$ actions is chosen a constant number of times $C$, yielding a set of $A \cdot C$ children. Sparse sampling is then run on each of the children with a recursion depth one less than the root's. Once the tree is fully created, the leaves are each assigned a value of zero. Then, starting at the leaves, the values are backed up and combined via the Bellman equation, giving the parents' values, until the root's value is determined. The total number of nodes in this search tree is $(AC)^d$, making the algorithm impractical to run in all but the most trivial of domains.

It is worth noting, however, that **Sparse Sampling** is best known as one of the first RL planning algorithms that can achieve high accuracy with high probability using an amount of computation that is not a function of the size of the state space[2]. Because of this attractive property, it makes sense to select it or one of its variants as the planner for the infinitely large belief-MDP. **Sparse Sampling** is the basis for a number of Monte-Carlo Tree Search (MCTS) algorithms, which are considerably faster in practice (Kocsis & Szepesvári, 2006; Walsh et al., 2010; Wang et al., 2005).

**Bayesian Sparse Sampling** (Wang et al., 2005) is a modification of **Sparse Sampling** that applies only in the Bayesian setting. Instead of a full tree expansion, **Bayesian Sparse Sampling** preferentially expands only promising parts of the tree by performing rollouts, or simulated trajectories, up to the specified depth. On a given rollout, the action for a particular node's belief-state is chosen by sampling a model from the posterior and solving this model exactly for the current "real" state. This action-selection strategy is myopic, but the algorithm can still achieve Bayes-optimal behavior in the limit because the method for computing the resulting policy is the same as in **Sparse Sampling**; it propagates values from the leaves back towards

---

[2]Assuming sampling from the model in constant time

**Input**: state $s$, max depth $d$, #trajectories $t$, MDP $M$
**Output**: estimated value for state $s$
**for** $t$ **times do**
  | FSSS-Rollout$(s, d, 0, M)$
$\hat{V}(s) \leftarrow \max_a U_d(s, a)$
**return** $\hat{V}(s)$

**Algorithm 1:** FSSS$(s, d, t, M)$

the root. This algorithm is limited by the exact-solve step within the inner loop, but also works in domains with continuous action spaces, unlike **Sparse Sampling**.

**Upper Confidence bounds on Trees** (Kocsis & Szepesvári, 2006), or **UCT**, is a recent Monte-Carlo tree search idea that has gained notice through good performance in computer Go (Gelly & Silver, 2008). Like **Bayesian Sparse Sampling**, **UCT** preferentially expands the search tree by performing rollouts. In addition to the value estimates for each node, which are computed by running backups backwards along trajectories, **UCT** maintains upper confidence bounds using the **UCB** algorithm (Auer et al., 2002). This algorithm is very aggressive and typically will under-explore.

**Forward Search Sparse Sampling** (Walsh et al., 2010), or **FSSS**, is the approach we adopt in this paper. It also preferentially expands the search tree through the use of rollouts. It is outlined in Algorithm 1. Unlike either **Bayesian Sparse Sampling** or **UCT**, it retains the attractive guarantees of the original **Sparse Sampling** algorithm. **FSSS** maintains hard upper and lower bounds on the values for each state and action so as to direct the rollouts; actions are chosen greedily according to the upper bound on the value, and the next state is chosen such that it is the most uncertain of the available candidates (according to the difference in its upper and lower bounds).

**FSSS** will find the action to take from a given state $s_0$, which will be the root of the search tree. The tree is expanded by running $t$ trajectories, or rollouts, of length $d$. There are theoretically justified ways to choose $t$ and $d$, but in practical applications they are knobs used to balance computational overhead and accuracy. To run a single rollout, the agent will call Algorithm 2, FSSS-Rollout$(s_0, d, 0, M)$. The values $U_d(s)$ and $L_d(s)$ are the upper and lower bounds on the value of the node for state $s$ at depth $d$, respectively. Each time a rollout is performed, the tree will be expanded. After at most $(AC)^d$ rollouts are finished (but often less in practice), **FSSS** will have expanded the tree as much as is possibly useful, and will agree with the action chosen by **Sparse Sampling**.

## 4 Bayesian FSSS

This paper's contribution, **Bayesian FSSS (BFS3)** is the application of **FSSS** to a belief-MDP and is outlined in Al-

**Input**: state $s$, max depth $d$, current depth $l$, MDP $M$
**if** *Terminal*$(s)$ **then**
    $U_d(s) = L_d(s) = 0$
    **return**
**if** $d = l$ **then**
    **return**
**if** $\neg Visited_d(s)$ **then**
    $Visited_d(s) \leftarrow$ true
    **foreach** $a \in A$ **do**
        $R_d(s,a), \text{Count}_d(s,a,s'), \text{Children}_d(s,a)$
            $\leftarrow 0, 0, \{\}$
        **for** $C$ *times* **do**
            $s', r \sim T_M(s,a), R_M(s,a)$
            $\text{Count}_d(s,a,s') \leftarrow \text{Count}_d(s,a,s') + 1$
            $\text{Children}_d(s,a) \leftarrow \text{Children}_d(s,a) \cup \{s'\}$
            $R_d(s,a) \leftarrow R_d(s,a) + r/C$
            **if** $\neg Visited_{d+1}(s')$ **then**
                $U_{d+1}(s'), L_{d+1}(s') = V_{\max}, V_{\min}$
    Bellman-backup$(s,d)$
$a \leftarrow \operatorname{argmax}_a U_d(s,a)$
$s' \leftarrow \operatorname{argmax}_{s'} (U_{d+1}(s') - L_{d+1}(s')) \cdot \text{Count}_d(s,a,s')$
FSSS-Rollout$(s', d, l+1, M)$
Bellman-backup$(s,d)$
**return**
        **Algorithm 2:** FSSS-Rollout$(s,d,l,M)$

**Input**: state $s$, history $h$, depth $d$, #trajectories $t$, MPD
      prior $\phi$
**Output**: action to take in state $s$
**if** $\langle s, h \rangle \in$ *solved-belief-states* **then**
    **return** $\pi(\langle s, h \rangle)$
**foreach** $a \in A$ **do**
    **for** $C$ *times* **do**
        $\langle s', h' \rangle, r \sim T\text{-}R_\phi(\langle s, h \rangle, a)$
        $q(a) \leftarrow q(a) + \frac{1}{C}[r + \gamma FSSS(\langle s', h' \rangle, d, t, M_\phi)]$
solved-belief-states $\leftarrow$ solved-belief-states $\bigcup \{\langle s, h \rangle\}$
$\pi(\langle s, h \rangle) \leftarrow \operatorname{argmax}_a q(a)$
**return** $\pi(\langle s, h \rangle)$
        **Algorithm 3:** BFS3$(s, h, d, t, \phi)$

gorithm 3. For some MDP prior $\phi(M)$, the joint transition and reward function $T\text{-}R_\phi$ is constructed such that

$$P(\langle s', h \cup (s,a,s',r)\rangle, r | \langle s, h \rangle, a) = \int_M P(s', r | s, a, M) \phi(M|h) dM.$$

Since, with **FSSS**, the next belief-states are only sampled and their likelihoods are never calculated, a simple generative process can be used:

$$M \sim \phi|h \quad (1)$$
$$s', r \sim T_M(s,a), R_M(s,a). \quad (2)$$

This process is used whenever **BFS3** or its subroutine **FSSS** sample a next-state and reward. The algorithm never holds on to an individual MDP after a single transition has been sampled from it. Also, note that whenever **FSSS** does a Bellman backup, that backup is done for a belief-state (since **FSSS** is acting on the belief-MDP).

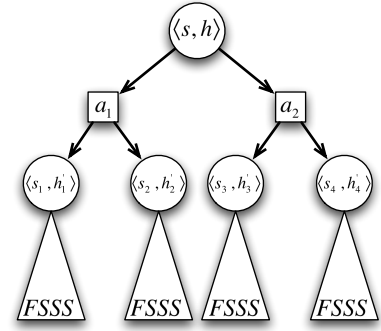

Figure 1: **BFS3** samples next-belief-states for each action $C$ times, and then runs **FSSS** on the resulting belief state, using the belief-MDP as a generative model. Every node the same distance from the root represents one of the possible worlds that the agent may experience, each with a different history and MDP posterior.

First, an MDP $M$ is sampled from the posterior $\phi|h$, and then the next state and reward are sampled from $M$. To reconstruct the resulting belief-state, we pack $s'$ with the new history made by augmenting $h$ with $(s, a, s', r)$, resulting in a transition from belief-state $\langle s, h \rangle$ to $\langle s', h \cup (s, a, s', r) \rangle$.

Figure 1 illustrates the **BFS3**'s process for each belief-state visited by the agent. In the future, the agent may find itself in one of the reachable belief-states in the search tree.

In many cases, the history $h$ can be summarized by a more compact sufficient statistic, and next-state posteriors can be sampled efficiently (for example, the **FDM** prior detailed in Section 2).

## 5 Near Bayes-optimal behavior

An agent achieves Bayes-optimal behavior (Wang et al., 2005; Duff & Barto, 1997) if it operates according to a policy that maximizes expected discounted cumulative reward with respect to some MDP prior $\phi$. Built into this policy are both explorative and exploitative actions. To the Bayes-optimal agent, exploration and exploitation are the same thing. An agent is near Bayes-optimal if it has approximately Bayes-optimal behavior for all but a number of steps that is polynomial with the parameters of the environment.

If one were to run **Sparse Sampling** on the belief-MDP out to the correct depth, it would return the approximately Bayes-optimal policy. Since **FSSS**, when given enough time, always agrees with **Sparse Sampling**, we know that as we give it more computational resources the resulting policy will approach approximate Bayes-optimality.

The interesting case is when computation time is limited, and **FSSS** cannot expand the full search tree that **Sparse Sampling** would discover. We shall consider computation time in terms of queries to the oracle, that is, times samples are drawn from the distribution described in Equations 1 and 2.

For a given value estimation, **FSSS** will query the oracle at most $t \cdot d \cdot A \cdot C$ times, where $t$ is the number of trajectories that will be simulated, $d$ is the maximum depth of any trajectory, $A$ is the number of actions available, and $C$ is the number of times each action is tried for a given node in the tree. On a step for which **BFS3** needs to plan, it will use **FSSS** to estimate the value for $C \cdot A$ different states, for a total of $t \cdot d \cdot A^2 \cdot C^2$ oracle queries.

To determine whether or not this number of queries is efficient, we can examine each of the factors. $A$ is provided by the domain. For a particular desired accuracy $\epsilon$, states beyond a certain depth have too small an impact on the value at the root (due to discounting). This depth $d$ grows (or shrinks) with $\log_\gamma \epsilon$.

The remaining factors, $t$ and $C$, are often considered to be free parameters. As $t$ and $C$ grow, so does the accuracy of the computed policy. Theory gives worst-case bounds for both of these values. In the worst case (combination lock (Li, 2009)), $t$ must be equal to the number of possible combinations, or $A^d$. In the worst case, $C$ must be so high as to paint a picture that is accurate for all $t \cdot d \cdot A$ estimated transition functions with high probability, simultaneously, resulting in a ludicrous number that can only be used in the most trivial of domains.

In practice, both $t$ and $C$ may be set to values significantly smaller than the worst case. Many domains are "deterministic with noise", meaning that only a few queries are necessary for each node—enough to be sure of the next-state distribution's mode—and $C$ can be quite low. Also, many domains have widely varying rewards that can help an MCTS planner decide which parts of the state space to consider first, allowing $t$ to be much more manageable.

Therefore, we shall operate on the theoretically dubious but practically reasonable premise that both $t = t_0$ and $C = C_0$ are small, but large enough that an agent using **FSSS** (equipped with the true model as its oracle) to plan will behave approximately optimally. We will express our bounds in terms of these parameters.

## 5.1 Proof of Near Bayes-optimality

We next lay out an argument for why **BFS3** has near Bayes-optimal behavior for discrete state and action space MDPs with known reward functions and known terminal states. In this context, a sufficient statistic for the belief-state is the pairing of the real state and the next-state histograms for each state-action pair.

### 5.1.1 Theorem statement

If: 1. **FSSS**, given the true MDP $m_0$ and some bound on computational resources, can provide accurate value estimates with high probability, and 2. the posterior next-state distribution for some state-action pair, given $N$ examples, will be an accurate estimate of $m_0$'s next-state distribution with high probability, then: with high probability, **BFS3** will behave Bayes-optimally for all but a polynomial number of steps.

### 5.1.2 Proof (sketch)

First, we will show that there is a belief-MDP, constructed from the prior $\phi$, whose optimal policy is the Bayes-optimal policy for $m_0 \sim \phi$. Then, we will show that **BFS3**, acting in the belief-MDP, will satisfy the three criteria required for PAC-MDP behavior (Kakade, 2003; Li, 2009) in that belief-MDP[3]. These criteria are: 1. accuracy, 2. bounded discoveries, and 3. optimism.

First, because of Condition 2 in our theorem statement, we know that once we have received $N$ examples of transitions from a state-action pair $(s, a)$, our estimate of the next-state distribution for that pair will be accurate. (This condition need not hold for degenerate priors, but it appears to hold quite broadly.)

Second, since we forget all additional transitions from state-action pairs for which we have seen $N$ examples, the number of possible state-histories that an agent can observe is bounded. Specifically, each time a transition from some state-action $(s, a)$ is observed, either no change will be made to the state-action's histogram (it already sums to $N$), or exactly one entry in the histogram will be incremented by 1. Since the histogram can be changed at most $N$ times, the total number of histories possible for an agent over the course of a single experiment is $N \cdot S \cdot A$ ($N$ histories for each state-action pair).

A discovery event, or one that potentially changes the MDP posterior, is an event that results in a change to the history. There are $N \cdot S \cdot A$ discoveries possible, since other transitions will be forgotten.

Third, **FSSS**$(s', d, t, M_\phi)$ is guaranteed to have an optimistic value estimate for belief-state $s'$ as $t$ (the number of trajectories), our bounded resource, grows smaller. We also know that, from Condition 1 of the theorem, $t$ is sufficient to find accurate estimates of $s'$ if all states in $s'$'s subtree have converged next-state posteriors. Simply put, if $s'$'s subtree has no unknown state-action pairs, then **FSSS**'s estimate of that state's value will be accurate. As a result, if **FSSS**'s estimate of a state's value is inaccurate, there must be something to learn about in $s'$'s subtree. **FSSS** guarantees that this inaccuracy will be optimistic.

---

[3]PAC-MDP behavior in the belief-MDP implies near Bayes-optimal behavior in the learning setting.

Also possible is that the value estimate of $s'$ is accurate *and* there are unknown states in its subtree. In this case, the agent can decide whether or not to visit that state fully informed of its value, and can take a Bayes-optimal action.

The PAC-MDP criteria direct the agent to areas of either high value or high uncertainty, managing the exploration/exploitation tradeoff. Because the agent will only go to areas of high uncertainty over areas of high reward a bounded number of times that grows linearly with the number of possible discovery events, we bound the number of sub-optimal steps taken over the lifetime of the agent.

For a formal proof of **BFS3**'s near Bayes-optimality, see the appendix (Asmuth & Littman, 2011).

The theoretical result given has practical implications. **BFS3**, in effect, has a *computational resources* knob. When the knob is set to allow unbounded computation, **BFS3** will make Bayes-optimal decisions. When the knob is tuned to only allow the minimal computation needed to solve a sampled MDP, **BFS3** will behave no worse than a PAC-MDP algorithm. Thus, the algorithm introduces a new kind of exploration/computation tradeoff that is not present in existing model-based RL approaches. The community has been grappling with this tradeoff at a high level by proposing algorithms at various points along the spectrum. Ours is the first algorithm that treats this tradeoff parametrically.

## 6 Experimentation

To demonstrate **BFS3**, we will show its performance in a number of domains, and show how the use of different priors can affect its performance. This flexibility with respect to using different priors is a compelling reason to use MCTS algorithms in general, and **BFS3** in particular, for model-based reinforcement learning.

When using the **FDM** prior with a small state space, **BEB** may be considered a better choice. Since **BEB** operates greedily according to a grounded MDP, rather than a belief-MDP, planning is made potentially much easier. This algorithm is limited, however, in that it requires a known reward function; it can only deal with uncertainty in the transition function.

**BFS3**, with the right prior, can handle unknown rewards. In many domains, there are only a few possible reward values. For example, many path-finding domains give a reward of $-1$ for all actions. Or, there is a reward for a particular outcome that can be achieved from multiple states: these states would share the same reward value. To represent this common structure in a generative model, the Dirichlet Process (DP) (MacEachern & Muller, 1998) may be used:

$$R_{s,a} \sim \text{DP}(\alpha, \text{Unif}(R_{\min}, R_{\max})).$$

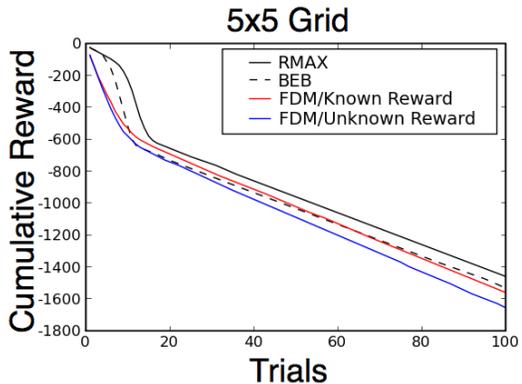

Figure 2: Here **BFS3/FDM**, with both known rewards and unknown rewards with a DP prior, is compared against the known-reward **BEB**, and **RMAX** (Brafman & Tennenholtz, 2002), a well known algorithm with PAC-MDP guarantees, with $M = 5$. Results are averaged over 40 runs.

Note that with this prior, rewards are deterministic, if unknown.

In Figure 2, **BFS3**, with the unknown reward prior, is shown to suffer no significant performance penalty compared to **BFS3** with the known-reward prior and to **BEB** (which also uses the known-reward prior). The domain is a $5 \times 5$ grid world, where the agent must find its way from one corner to the other. The agent can choose to go north, east, south or west, and with probability $0.2$ it will go in a direction perpendicular to the one intended.

Along with **FDM**, we introduce the **Factored-Object** prior, which describes factored MDPs in which the state features are broken up into a number of independent yet identical objects. The action also has two features: the first indicates which object is being acted upon, and the second indicates which action is being performed. **Factored-Object** essentially has a single **FDM** posterior, which it applies to each object in the state simultaneously, sharing both information (for faster convergence) and memory.

For a single object, **Factored-Object** and **FDM** are the same. For two objects, **FDM** has to learn separately how a particular action affects a particular object for every possible configuration of objects—for a different state of an object not being operated on, **FDM** must re-learn how the original object is affected. **Factored-Object** allows the agent to learn about multiple objects at the same time: it knows that a given action affects object 1 in the same way it affects object 2, and generalizes appropriately.

The Paint/Polish world (Walsh et al., 2009) provides a situation where the simple and convenient **FDM** prior is insufficient. The size of the state-space grows exponentially with the number of cans to paint (each of which introduces four binary features). Figure 3 shows the results with a single can (and $2^4$ states) and the results with two cans (and $2^{4 \cdot 2}$ states). If the number of cans is increased to four, **FDM** cannot find a good policy in a reasonable amount of time while **Factored-Object** can do so in around ten trials.

**BFS3** can also be used to apply Bayesian modeling to POMDPs. Wumpus World (Russell & Norvig, 1994) is based a classic computer game in which an agent wanders through a $4 \times 4$ maze filled with fog, making it impossible to see past its current cell. Even though the agent cannot see, it can feel a breeze if there is a pit in an adjacent cell, and smell a stench if there is a Wumpus[4] near-by. If the agent falls into a pit, it will remain there forever. If the agent runs into the Wumpus, it is eaten. If the agent shoots its one arrow in the direction of the Wumpus, the Wumpus is slain. If the arrow misses the Wumpus, the trial ends and presumably the agent goes home. We replicate the dynamics presented in detail by Sorg et al. (2010).

Wumpus World is based on a deterministic process, but since the agent only knows attributes of cells that it has visited, it appears stochastic. The prior over different possible mazes is known and given to the agent, and from this it can infer the correct posterior distribution over what happens when it performs a particular action in a particular belief-state.

We ran **BFS3** on Wumpus World with a search depth of 15, and varied the number of trajectories per step. Agents with 500, 1000, and 5000 trajectories per step averaged 0.267, 0.358 and 0.499 cumulative reward, respectively. Averages were taken over 1000 attempts. We compare this to a variance-based reward bonus strategy (Sorg et al., 2010) which, when tuned, averaged 0.508.

That **BFS3** performs better in Wumpus World as the computation budget is increased supports our argument that the algorithm has a *computational resources* knob which, when tuned higher, causes the agent's behavior to get closer to being Bayes-optimal at the cost of decision-making speed.

## 7 Concluding Remarks

The use of Bayesian methods with flexible priors has been a boon to machine learning and artificial intelligence. Similarly, the use of model priors in Bayesian reinforcement learning is a powerful tool for incorporating prior knowledge into an algorithm in a principled way. However, most algorithms that make use of any prior use only the Flat Dirichlet Multinomial (**FDM**). While appropriate for many situations, this prior does not provide for any generaliza-tion across states or for continuous state spaces severely limiting its applicability.

[4] A Wumpus is a monster that eats RL agents.

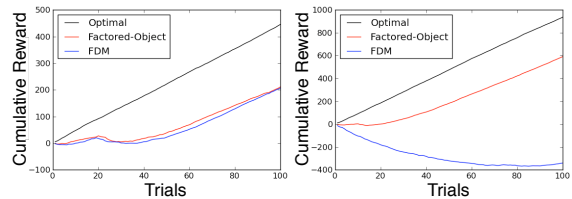

Figure 3: **BFS3** with **FDM** and **Factored-Object** priors. **Left:** Paint/Polish with 1 can. **FDM** and **Factored-Object** are identical for one object, and give the same performance. **Right:** Paint/Polish with 2 cans. **Factored-Object** outperforms **FDM** because it is better able to generalize. Results are averaged over 40 runs.

**Cluster** is an interesting prior that can be used instead of **FDM**. In degenerate cases, it will converge to the same point as **FDM**, but very often it will detect that groups of states share dynamics and can be clustered together. Algorithms that limit themselves to **FDM** cannot make use of this or any other generalizing prior.

Though limited in other ways, **FDM** does have at least one property that makes it more attractive than many other prior distributions: it makes posterior sampling trivial. Especially for MCTS approaches, posterior sampling needs to be fast, as it is done often. The aforementioned **Cluster** prior is difficult to use with MCTS strategies because sampling from its posterior involves Markov chain Monte-Carlo techniques (Neal, 2000), and requires a great deal of computation.

**BOSS** (Asmuth et al., 2009) can make use of flexible priors, and uses sample redundancy and posterior variance to ensure constant optimism, and is PAC-MDP. However, it is very conservative and will often over-explore. Like **BOSS**, the variance-based reward bonus approach (Sorg et al., 2010) draws upon posterior variance of any prior to encourage exploration to unknown states, but does so in a different way that gives it near Bayes-optimality.

The algorithm we have introduced, **BFS3**, has the best of both worlds: it can make use of a wide variety of priors, and it is near Bayes-optimal. Its weakness is computation. For some prior distributions (**Cluster**), posterior samples can be expensive, and for others a large branching factor $C$ must be used. Both increase the computational power required to ensure good behavior. Despite this weakness the great flexibility it offers, in terms of domains and of priors, make it an attractive application of Bayesian techniques for reinforcement learning.

PAC-MDP and near Bayes-optimal differ in important ways. Near Bayes-optimal is a claim made in the context of some prior distribution, where PAC-MDP is a claim made in the context of any possible underlying model. Algorithms that are PAC-MDP will therefore over-explore, while near Bayes-optimal algorithms will explore just the right amount, according to the prior.

The exploration/exploitation dilemma is central to the RL community. PAC-MDP and Bayes-optimal guarantees represent two important approaches to this problem. PAC-MDP algorithms address the issue by spending a budget of exploration steps toward improving their models to the point at which they are accurate. Bayes-optimal algorithms address the issue by planning in belief-space, turning the exploration/exploitation dilemma into just exploitation: learning is planning.